# Hidden Markov Chains, Entropic Forward-Backward, and Part-Of-Speech Tagging

E. Azeraf, E. Monfrini, E. Vignon, and W. Pieczynski

*Abstract*— **The ability to take into account the characteristics - also called features - of observations is essential in Natural Language Processing (NLP) problems. Hidden Markov Chain (HMC) model associated with classic Forward-Backward probabilities cannot handle arbitrary features like prefixes or suffixes of any size, except with an independence condition. For twenty years, this default has encouraged the development of other sequential models, starting with the Maximum Entropy Markov Model (MEMM), which elegantly integrates arbitrary features. More generally, it led to neglect HMC for NLP. In this paper, we show that the problem is not due to HMC itself, but to the way its restoration algorithms are computed. We present a new way of computing HMC based restorations using original Entropic Forward and Entropic Backward (EFB) probabilities. Our method allows taking into account features in the HMC framework in the same way as in the MEMM framework. We illustrate the efficiency of HMC using EFB in Part-Of-Speech Tagging, showing its superiority over MEMM based restoration. We also specify, as a perspective, how HMCs with EFB might appear as an alternative to Recurrent Neural Networks to treat sequential data with a deep architecture.**

## I. Introduction

HIDDEN Markov Chain (HMC) is a very popular model, used in innumerable applications [1][2][3][4][5]. For about sixty years, it is very appreciated for its simplicity, interpretability, robustness, and learning speed. However, in some applications, its simplicity poses problems. In particular, they were shown to be inadequate to deal with text segmentation tasks, and they were abandoned in favor of the Maximum Entropy Markov Model [6] (MEMM). Indeed, for these tasks, the ability to use arbitrary information of a word as its suffixes, its prefixes, if its first letter is up, etc. is mandatory to achieve relevant results [7]. HMC based restoration algorithms, Viterbi [8] and Forward-Backward [3], are unable to take into account these characteristics - also called features - except with an independence constraint that turns out to be too strong for Natural Language Processing (NLP) tasks.

This default of the HMC's based restoration algorithms has motivated the development of new models adapted to use these pieces of information in textual data. The MEMM [6] was specially developed to overstep this "feature" problem of HMC. We can also cite Conditional Random Fields [9], Recurrent Neural Network (RNN) [10] [11], among others.

In this paper, we show that the "feature" problem above is not due to the model but concerns how the classic restoration algorithms are computed. We propose a new way to calculate the Maximum Posterior Mode (MPM) restoration algorithm using original "entropic forward-backward" probabilities, which allows HMC to handle arbitrary features, as MEMM does. Moreover, we show the superiority of HMC's based MPM over the MEMM's based one for the Part-Of-Speech (POS) tagging, one of the main NLP segmentation tasks.

As a perspective, we briefly discuss the potential of this new algorithm to allow HMCs to deal with "deep" sequential data in the framework of neural networks, a task mainly done using RNNs and their extensions until now.

The organization of the paper is the following. We recall the definition of HMC and the Forward-Backward algorithm in the next section. We also specify the problems arising when applying HMC to any NLP task, as POS tagging. Section III is devoted to the new Entropic Forward-Backward (EFB) probabilities and their application to MPM computation. We recall MEMM and RNN models in section IV, and section V is devoted to experiments comparing MEMM and HMC with EFB. Conclusions and perspectives lie in section VI.

## II. Hidden Markov Chains

Let $X_{1:T} = (X_1,...,X_T)$ be a hidden stochastic chain, and $Y_{1:T} = (Y_1,...,Y_T)$ an observed one. For $t = 1, ..., T$, $X_t$ takes its values in the set of states, or labels, $\Lambda_X = \{\lambda_1,...,\lambda_N\}$, and $Y_t$ takes its values in $\Omega_Y = \{\omega_1,...,\omega_M\}$. Realizations of random variables will be denoted with minus letters. The couple $(X_{1:T}, Y_{1:T})$ is a HMC if, for $t = 1, ..., T$:

$$p(x_{1:T}) = p(x_1)p(x_2|x_1) ... p(x_T|x_{T-1}); \quad (1)$$

$$p(y_{1:T}|x_{1:T}) = p(y_1|x_1) ... p(y_T|x_{1:T}); \text{ and} \quad (2)$$

$$p(y_t|x_{1:T}) = p(y_t|x_t). \quad (3)$$

E. Azeraf and E. Vignon are with IBM GBS France, Watson Department, 17 Avenue de l'Europe, 92275 Bois-Colombes, France (e-mail: {elie.azeraf, emmanuel.vignon}@ibm.com). E. Azeraf, E. Monfrini, and W. Pieczynski are with Institut Polytechnique de Paris, Telecom Sudparis, SAMOVAR, 9, rue Charles Fourier, 91 000 Evry, France. (e-mail: {Elie.Azeraf, Emmanuel.Monfrini, Wojciech.Pieczynski}@telecom-sudparis.eu).

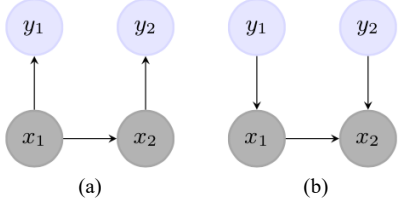

Fig 1. Probabilistic oriented dependence graphs of HMC (a), and MEMM (b)

This is equivalent to assume that the distribution $p(x_{1:T}, y_{1:T})$ is of the form

$$p(x_{1:T}, y_{1:T}) = p(x_1)p(x_2|x_1)...p(x_T|x_{T-1})p(y_1|x_1)....p(y_T|x_T). \quad (4)$$

The oriented probabilistic dependence graph of HMC is presented in Figure 1, (a).

Given an observed sequence $Y_{1:T} = y_{1:T}$, there are mainly two Bayesian methods to estimate the realization of the hidden chain $X_{1:T}$: the MPM and the Maximum A Posteriori (MAP). The solution of MAP is given with the Viterbi algorithm [8], and the solution of MPM follows directly from the Forward-Backward [3] (FB) algorithm. In this paper, we concentrate on MPM.

We consider stationary HMCs, and we adopt the following notations. For each $(\lambda_i, \lambda_j) \in \Lambda_X^2$, $y \in \Omega_Y$:

$$\pi_i = P(X_t = \lambda_i), \text{ for } t = 1, ..., T; \quad ; \quad (5)$$

$$a_{ij} = P(X_{t+1} = \lambda_j | X_t = \lambda_i), \text{ for } t = 1, ..., T-1; \quad (6)$$

$$b_i(y) = P(Y_t = y | X_t = \lambda_i), \text{ for } t = 1, ..., T. \quad (7)$$

In a supervised framework, we estimate these parameters with maximum likelihood, which consists of evaluating frequencies of the different patterns [12][6]. We have to browse one time the training set to learn the parameters. To compute MPM, one classically uses Forward and Backward probabilities.

Forward probabilities are defined as

$$\alpha_t(i) = P(X_t = \lambda_i, y_{1:t}), \quad (8)$$

and Backward probabilities are defined as

$$\beta_t(i) = P(y_{t+1:T} | X_t = \lambda_i). \quad (9)$$

Then, classically:

$$P(X_t = \lambda_i | y_{1:T}) = \frac{\alpha_t(i)\beta_t(i)}{\sum_{\lambda_j \in \Omega_X} \alpha_t(j)\beta_t(j)}, \quad (10)$$

which leads to MPM, denoted with $\hat{s}_{MPM}$. We have

$$[\hat{s}_{MPM}(y_{1:T}) = \hat{x}_{1:T}] \Leftrightarrow$$
$$[\forall t \in \{1,...,T\}, P(X_t = \hat{x}_t | y_{1:T}) = \sup_{\lambda_i \in \Omega_X} P(X_t = \lambda_i | y_{1:T})] \quad (11)$$

The interest of HMC based MPM is that the following Forward and Backward recursions can compute forward and backward probabilities:
For $y_{1:T}$, $t = 1, ..., T-1$, $\lambda_i \in \Omega_X$:

$$\alpha_1(i) = \pi_i b_i(y_1), \alpha_{t+1}(i) = b_i(y_{t+1}) \sum_{j=1}^{N} \alpha_t(j) a_{ji}, \quad (12)$$

and

$$\beta_T(i) = 1, \beta_t(i) = \sum_{j=1}^{N} \beta_{t+1}(j) a_{ij} b_j(y_{t+1}). \quad (13)$$

When using HMC with Forward-Backward or Viterbi in NLP tasks, one has to face the following problem. Let us consider, as an example, the POS tagging. $y_1, ..., y_T$ are words, $\Omega_Y$ is the set of all possible words, and $x_1, ..., x_T$ are grammatical labels, like Adjective, Verb, Noun, Determinant..., and thus $\Lambda_X$ is the set of such labels.

We have to estimate parameters ($\pi_i$, $a_{ij}$, $b_i(y)$), for $\lambda_i, \lambda_j \in \Lambda_X$, $y \in \Omega_Y$, from a training set $D$ composed of sentences of labelled words, and use them to label new sentences. For a new sentence $y_{1:T}$, if the word $y_t$ never appears in the training set, we have $b_i(y_t) = 0$ for each $\lambda_i \in \Lambda_X$. With possibly many words not in the training set, this effect can lead to many errors for these "unknown" words. One way to solve this problem is to use an approximation of $b_i(y_t)$ thanks to the features of an observation. Features can be prefixes, suffixes, length of the word, etc. It implies that a word $y_t$ becomes a vector $(y_t^1,...,y_t^K)$ with $K$ features, and $b_i(y_t)$ becomes $P(Y_t^1 = y_t^1,...,Y_t^K = y_t^K | X_t = \lambda_i)$. If this new modelization is tractable for small values of $K$, it quickly leads to assume that, knowing $X_t = \lambda_i$, features are independent and to set

$$P(Y_t^1 = y_t^1,...,Y_t^K = y_t^K | X_t = \lambda_i) = P(Y_t^1 = y_t^1 | X_t = \lambda_i)...P(Y_t^K = y_t^K | X_t = \lambda_i). \quad (14)$$

This independence hypothesis can be made with success for many applications but is irrelevant for NLP. Some works propose approximation methods to handle some features for NLP tasks [12], but the problem remains since we cannot choose arbitrary features (in general, these works only select the suffixes and few other hand-crafted features).
We are going to see that it is possible to work around this problem by modifying the way of computing the MPM solution. FB probabilities (8)-(9) will be replaced with new "Entropic" FB, which will no longer use $b_i(y_t)$ at the origin of problems, but which will still allow computation of MPM, in a similar way that MEMMs do.

III. ENTROPIC FORWARD-BACKWARD (EFB) CALCULATION

Let us notice that classic Forward and Backward probabilities have been introduced with (8)-(9), and then (12)-(13) have been derived as properties. However, the converse is possible: they may be defined with (12)-(13) and then (8)-(9) would appear as properties. We will use this method to define the proposed EFB probabilities.

For each $y_{1:T}$, $t = 1, ..., T$, $\lambda_i \in \Lambda_X$, let:

$$L_{y_t}(i) = P(X_t = \lambda_i | y_t) \quad (15)$$

*Definition*
Entropic Forward (EF) probabilities are defined with the following forward recursion:

For each $y_{1:T}$, $t = 1, \ldots, T-1$, $\lambda_i \in \Lambda_X$,

$$\alpha_1^E(i) = L_{y_1}(i); \quad \alpha_{t+1}^E(i) = \frac{L_{y_{t+1}}(i)}{\pi_i} \sum_{j=1}^N \alpha_t^E(j) a_{ji}, \quad (16)$$

Entropic Backward (EB) probabilities are defined with the following backward recursion:
For each $y_{1:T}$, $t = 1, \ldots, T$, $\lambda_i \in \Lambda_X$,

$$\beta_T^E(i) = 1; \quad \beta_t^E(i) = \sum_{j=1}^N \frac{L_{y_{t+1}}(j)}{\pi_j} \beta_{t+1}^E(j) a_{ij}, \quad (17)$$

We can state:

*Proposition*

Let $\alpha_t^E(i)$ and $\beta_t^E(i)$ be Entropic Forward and Entropic Backward probabilities, respectively. Posterior marginal distributions verify

$$P(X_t = \lambda_i | y_{1:T}) = \frac{\alpha_t^E(i) \beta_t^E(i)}{\sum_{j=1}^N \alpha_t^E(j) \beta_t^E(j)}. \quad (18)$$

*Proof.*
Let us show that for each $\lambda_i \in \Omega_X$, $t = 1, \ldots, T$,

$$\alpha_t^E(i) = \frac{\alpha_t(i)}{p(y_1) \ldots p(y_t)}; \quad (19)$$

$$\beta_t^E(i) = \frac{\beta_t(i)}{p(y_{t+1}) \ldots p(y_T)}. \quad (20)$$

To show (19) we use forward induction. (19) is true for $t = 1$; let us assume it is true for $t > 1$. As $P(X_t = \lambda_i, Y_t = y_t) = L_{y_{t+1}}(i) p(y_{t+1}) = \pi_i b_i(y_{t+1})$, we have $L_{y_{t+1}}(i) = \frac{\pi_i b_i(y_{t+1})}{p(y_{t+1})}$. Reporting it into (16) and using (19) gives

$$\alpha_{t+1}^E(i) = \frac{b_i(y_{t+1})}{p(y_{t+1})} \sum_{j=1}^N \alpha_t^E(j) a_{ji} =$$

$$\frac{b_i(y_{t+1})}{p(y_{t+1})} \sum_{j=1}^N \frac{\alpha_t(j)}{p(y_1) \ldots p(y_t)} a_{ji} = \frac{\alpha_{t+1}(i)}{p(y_1) \ldots p(y_t) p(y_{t+1})},$$

and thus (12) is true for $t+1$, which shows (12) for each $t = 1, \ldots, T$.
Similarly, (20) is proved using backward induction. It is true for $t = T$; let us assume that it is true for $t+1 < T+1$.
Reporting $L_{y_{t+1}}(j) = \frac{\pi_j b_j(y_{t+1})}{p(y_{t+1})}$ into (17) and using (20) gives

$$\beta_t^E(i) = \frac{1}{p(y_{t+1})} \sum_{j=1}^N b_j(y_{t+1}) \beta_{t+1}^E(j) a_{ij} =$$

$$\frac{1}{p(y_{t+1}) \ldots p(y_T)} \sum_{j=1}^N b_j(y_{t+1}) \beta_{t+1}(t) a_{ij} = \frac{\beta_i(t)}{p(y_{t+1}) \ldots p(y_T)}$$

and thus (20) is true for $t$, which shows (20) for each $t = 1, \ldots,$ $T-1$. (19) and (20) imply $\alpha_t^E(i) \beta_t^E(i) = \frac{\alpha_t(i) \beta_t(i)}{p(y_1) \ldots p(y_T)}$, and thus $\frac{\alpha_t^E(i) \beta_t^E(i)}{\sum_{j=1}^N \alpha_t^E(j) \beta_t^E(j)} = \frac{\alpha_t(i) \beta_t(i)}{\sum_{j=1}^N \alpha_t(j) \beta_t(j)} = P(X_t = i | y_{1:T})$,

which ends the proof.

Finally, MPM can be computed using $L_{y_t}(i)$ instead of $b_i(y_t)$. Moreover, one can model $L_{y_t}(i)$ by any discriminative model [13] as logistic regression [14], feedforward neural networks [11], decision tree [15], etc. These models do not suffer from any constraint on features of observations. Therefore, the new way to compute MPM with Entropic Forward-Backward recursions above allow using arbitrary features for HMC's based restorations.

## IV. MAXIMUM ENTROPY MARKOV MODELS AND RECURRENT NEURAL NETWORKS

### A. Maximum Entropy Markov Models (MEMMs)

The MEMM has been created to handle arbitrary features for text segmentation, focusing on the inability of the algorithms of HMC to do that. The probabilistic graph of the MEMM is represented in Figure 1, (b). This discriminative model makes the following assumption.
For each $t = 1, \ldots, T-1$,

$$p(x_{t+1} | x_{1:t}, y_{1:t}) = p(x_{t+1} | x_t, y_t) \quad (21)$$

Therefore, the distribution of $(X_{1:T}, Y_{1:T})$ is of the form

$$p(x_{1:T} | y_{1:T}) = p(x_1 | y_1) p(x_2 | x_1, y_2) \ldots p(x_T | x_{T-1}, y_T) \quad (22)$$

In MEMM, variables $(X_t, Y_{1:t})$ and $Y_{t+1:T}$ are independent for each $t = 1, \ldots, T-1$ [16], so that the posterior marginal distributions verify

$$p(x_t | y_{1:T}) = p(x_t | y_{1:t}) \quad (23)$$

This implies that MPM uses only "forward" probabilities.
Let us set

$$\alpha_t^M(i) = P(X_t = \lambda_i | Y_{1:t}) \quad (24)$$

Let $L_{y_t}(i) = P(X_t = \lambda_i | y_t)$ given with (15), and let

$$L_{y_{t+1}}^1(l, i) = P(X_{t+1} = \lambda_i | X_t = \lambda_l, y_{t+1}) \quad (25)$$

We have the following forward recursion:

$$\alpha_1^M(i) = L_{y_1}(i),$$

$$\alpha_{t+1}^M(i) = \sum_{j=1}^N \alpha_t^M(j) L_{y_{t+1}}^1(l, i). \quad (26)$$

The main difference of the MPM restoration between MEMM and HMC resides thus in the fact that for each $t = 1, \ldots, T$, in MEMM, $X_t$ is estimated using only the observations from $1$ to $t$. In contrast, HMC considers all the observations for restoration thanks to the backward functions. Therefore, we can expect that HMC will have better results than MEMM if the features are handled in the same way.

## B. Recurrent Neural Network (RNN) model

The RNN is a model allowing the use of neural networks with sequential data with a "deep" architecture. We consider the classic RNN and not stacked one or the other extensions. Its computational graph is presented in Figure 2, (b). The RNN has three layers: the observation layer $y$, the hidden layer $h$, and the output of interest $x$.

In the RNN, the functions used are feedforward neural networks. Given the sequence $y_{1:T}$, we have for $t = 1, \ldots, T-1$:

$$h_{t+1} = f^1(y_{t+1}, h_t); \quad (27)$$

$$x_{t+1} = f^2(h_{t+1}), \quad (28)$$

with $f^1$ and $f^2$ modeled with neural networks. The parameters of the RNN are learned with the backpropagation algorithm [17]. We can notice similarities between the RNN and the MEMM. Indeed, MEMM is a RNN if $L$ and $L^1$ are modeled by feedforward neural networks. In the same idea, a RNN is a MEMM model if $f^2$ is the identity function. As MEMM, the classic RNN uses, for $t = 1, \ldots, T$, only observations from 1 to $t$ to predict $x_t$.

## V. EXPERIMENTS

We apply our algorithm to the POS Tagging. Let us recall that POS Tagging consists of labelling each word of a sentence with its grammatical function - Adjective, Verb, Noun, Determinant... For example, the sentence $y_{1:8} =$ *(Batman, is, the, main, vigilante, of, Gotham, .)* has the labels $x_{1:8} =$ *(PROPN, VERB, DET, ADJ, NOUN, DET, NOUN, PUNCT)*.

We compare the results of MEMM's based MPM with those of HMC's with Entropic Forward-Backward (HMC-EFB) based one. The discriminative functions $L$ (15) and $L^1$ (25) are modeled with logistic regression [14], trained with Stochastic Gradient Descent algorithm [18] [19]. We use the datasets CoNLL 2000 [20], CoNLL 2003 [21], and Universal Dependencies (UD) English [22]*, and the universal tagset [24] for our experiments.

The two models based MPM are coded in python by our own with the help of the PyTorch [25] library for gradient computation. The results of the experiments are in Table I. We present the percent of accuracy error for known words (KW), unknown words (UW), and the global dataset.

For each dataset, we consider the three following variants of MEMM and HMC-EFB's based MPMs:
- NF: no features are taken, only the words;
- LF$_1$, the first lot of features: the word itself, its suffixes and prefixes of size 3 and 2, if it is in the first position of the sentence and if it has its first letter up;
- LF$_2$, the second lot of features: with the same features of LF$_1$, to which we add suffixes and prefixes of size 5 and 4, if the word has a digit, if the word has a hyphen.

*All these datasets are freely available. CoNLL 2003 upon request at https:/www.clips.uantwerpen.be/conll2003/ner/, UD English is available on https:/universaldependencies.org/#language-, and CoNLL 2000 with the NLTK [23] library with python.

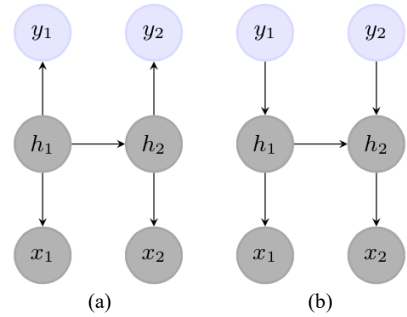

Fig. 2. Probabilistic oriented dependence graphs of TMC (a), and computational graph of RNN (b).

For each dataset, for each lot of features, HMC-EFB has globally better results than the MEMM. The gains can achieve 25%, with better performance in each case of unknown words. We also test the approximation methods to handle features as in [12], but the results of these methods were far from the ones of HMC-EFB in every case, and in some cases some features were impossible to be taken into account.

TABLE I: POS Tagging results (Known Words / Unknown Words / Global) for MEMM and HMC-EFB

|  | MEMM | HMC-EFB |
|---|---|---|
| **CoNLL 2000** |  |  |
| • NF | 2.21%/51.09%/5.62% | **2.07%/39.61%/4.69%** |
| • LF$_1$ | 2.21%/25.41%/3.83% | **2.06%/15.47%/2.99%** |
| • LF$_2$ | 2.20%/21.02%/3.51% | **2.06%/14.20%/2.90%** |
| **CoNLL 2003** |  |  |
| • NF | **3.82%/51.12%/9.11%** | 4.12%/48.23%/9.06% |
| • LF$_1$ | 3.81%/17.48%/5.35% | **4.04%/13.20%/5.07%** |
| • LF$_2$ | 3.81%/15.40%/5.11% | **4.02%/13.22%/5.04%** |
| **UD English** |  |  |
| • NF | 8.79%/59.60%/13.04% | **6.24%/53.69%/10.21%** |
| • LF$_1$ | 8.70%/35.49%/10.94% | **6.11%/28.87%/8.01%** |
| • LF$_2$ | 8.70%/31.40%/10.60% | **6.10%/27.54%/7.89%** |

## VI. HMC IN THE FOOTSTEPS OF RNN AS PERSPECTIVES

MEMM is a particular RNN when its functions are neural networks, and RNN is a very prevalent and intensively used model for NLP applications [10] [11] [25]-[31]. Otherwise, considering an extension of HMCs to "HMCs with hidden layers", similar to extensions of MEMMs to RNNs, we obtain a particular "triplet" Markov chain (TMC) [32]. Its probabilistic dependence graph, given in Figure 2, (a), looks like the computational graph of RNN in Figure 2, (b). TMCs have been successfully applied to different problems [33]-[42]. Once associated with EFB, and with parameters modeled with feedforward neural networks functions, they are likely to provide processing methods competing with RNNs based ones. In other words, TMCs with EFB extend HMCs with EFB, RNNs extend MEMMs, and HMCs with EFB outperform MEMMs. Therefore, we conjecture that in some NLP problems, TMCs with EFB would be competing with respect to RNNs. However, such extensions are not immediate, as learning is much more difficult because of hidden layers.